\documentclass[conference,a4paper]{IEEEtran}
\IEEEoverridecommandlockouts

\usepackage[hidelinks]{hyperref}
\usepackage[cmex10]{amsmath}
\usepackage{amssymb,amsfonts}
\usepackage{dblfloatfix}

\usepackage[ruled,vlined]{algorithm2e}
\usepackage{graphicx}
\graphicspath{{Figures/PDF/}{Figures/PNG/}}

\usepackage{booktabs}
\usepackage{multirow}
\usepackage{siunitx}
\usepackage[numbers,compress]{natbib}
\usepackage{texnames}
\usepackage{bm,bbm}
\usepackage{orcidlink}

\hypersetup{
	colorlinks=true,
	linkcolor=blue,
	citecolor=blue,
	urlcolor=red,
}

\begin{document}

\title{
	\LARGE\uppercase{Task-Driven Prompt Learning: A Joint Framework for Multi-modal Cloud Removal and Segmentation}
	\thanks{This work was supported in part by the National Natural Science Foundation of China under Grants 42471414 and 42471504.}
}

\author{
	\IEEEauthorblockN{Zaiyan Zhang \orcidlink{0009-0001-7376-7101}}
	\IEEEauthorblockA{
		\textit{Wuhan University}\\
		430079 Wuhan, China\\
		zzaiyan@whu.edu.cn}
	\and
	\IEEEauthorblockN{Jie Li \orcidlink{0000-0002-4063-9381}}
	\IEEEauthorblockA{
		\textit{Wuhan University}\\
		430079 Wuhan, China\\
		jli89@sgg.whu.edu.cn}
	\and
	\IEEEauthorblockN{Shaowei Shi \orcidlink{0009-0002-7694-9901}}
	\IEEEauthorblockA{
		\textit{Wuhan University}\\
		430079 Wuhan, China\\
		2017301610139@whu.edu.cn}
	\and
	\IEEEauthorblockN{Qiangqiang Yuan \orcidlink{0000-0001-7140-2224}}
	\IEEEauthorblockA{
		\textit{Wuhan University}\\
		430079 Wuhan, China\\
		qqyuan@sgg.whu.edu.cn}
}

\maketitle

\begin{abstract}
	Optical remote sensing imagery is indispensable for Earth observation, yet persistent cloud occlusion limits its downstream utility. Most cloud removal (CR) methods are optimized for low-level fidelity and can over-smooth textures and boundaries that are critical for analysis-ready data (ARD), leading to a mismatch between visually plausible restoration and semantic utility. To bridge this gap, we propose TDP-CR, a task-driven multimodal framework that jointly performs cloud removal and land-cover segmentation. Central to our approach is a Prompt-Guided Fusion (PGF) mechanism, which utilizes a learnable degradation prompt to encode cloud thickness and spatial uncertainty. By combining global channel context with local prompt-conditioned spatial bias, PGF adaptively integrates Synthetic Aperture Radar (SAR) information only where optical data is corrupted. We further introduce a parameter-efficient two-phase training strategy that decouples reconstruction and semantic representation learning. Experiments on the LuojiaSET-OSFCR dataset demonstrate the superiority of our framework: TDP-CR surpasses heavy state-of-the-art baselines by 0.18 dB in PSNR while using only 15\% of the parameters, and achieves a 1.4\% improvement in mIoU consistently against multi-task competitors, effectively delivering analysis-ready data.
\end{abstract}

\section{Introduction}

Optical remote sensing underpins a wide range of Earth observation applications, yet its availability and reliability are severely affected by clouds and cloud shadows \cite{king2013spatial}. Cloud removal (CR) aims to recover clear-sky imagery from cloudy observations, often with the help of auxiliary modalities such as synthetic aperture radar (SAR), which is insensitive to weather conditions \cite{shen2015missing}. Despite rapid progress, most CR models are still formulated and evaluated as low-level image restoration \cite{zhang2025multi, zhang2026ecrformer}, prioritizing pixel fidelity metrics.

However, remote sensing products are ultimately consumed by downstream analysis pipelines (e.g., Land Cover Segmentation, LCS), where edge integrity, texture, and small-object structures dominate semantic performance. This creates a persistent mismatch: a visually smooth reconstruction with high PSNR may suppress class-discriminative details and degrade segmentation/recognition, deviating from the practical goal of delivering \emph{analysis-ready data (ARD)}. This calls for a shift from \emph{visual restoration} to \emph{semantic restoration}.

To this end, we propose \textbf{TDP-CR} (Task-Driven Prompting for Cloud Removal), a joint framework for \emph{multimodal CR and LCS}. Our model uses decoupled encoders for cloudy optical and SAR inputs, and fuses features through a novel \textbf{Prompt-Guided Fusion (PGF)} block that is explicitly conditioned on a learnable degradation prompt. The prompt map indicates \emph{where} and \emph{to what extent} the optical stream is corrupted, enabling the network to selectively borrow reliable SAR cues without introducing heavy attention mechanism.

Our main contributions are summarized as follows:
\begin{itemize}
	\item \textbf{Task-driven Framework:} We bridge image processing and downstream analysis by coupling cloud removal with segmentation, prioritizing semantic utility over discrete pixel fidelity.
	\item \textbf{Prompt-Guided Fusion:} We propose a lightweight PGF module that leverages spatially adaptable degradation prompts to guide optical-SAR integration, ensuring robustness against varying cloud thickness.
	\item \textbf{Parameter-Efficient Fine-Tuning:} We introduce a two-phase strategy that decouples reconstruction features from semantic refinement, balancing generalizability with task-specific adaptation.
\end{itemize}

\section{Methodology}\label{sec:method}

\subsection{Overview}
\begin{figure}[t]
	\centering
	\includegraphics[width=\linewidth]{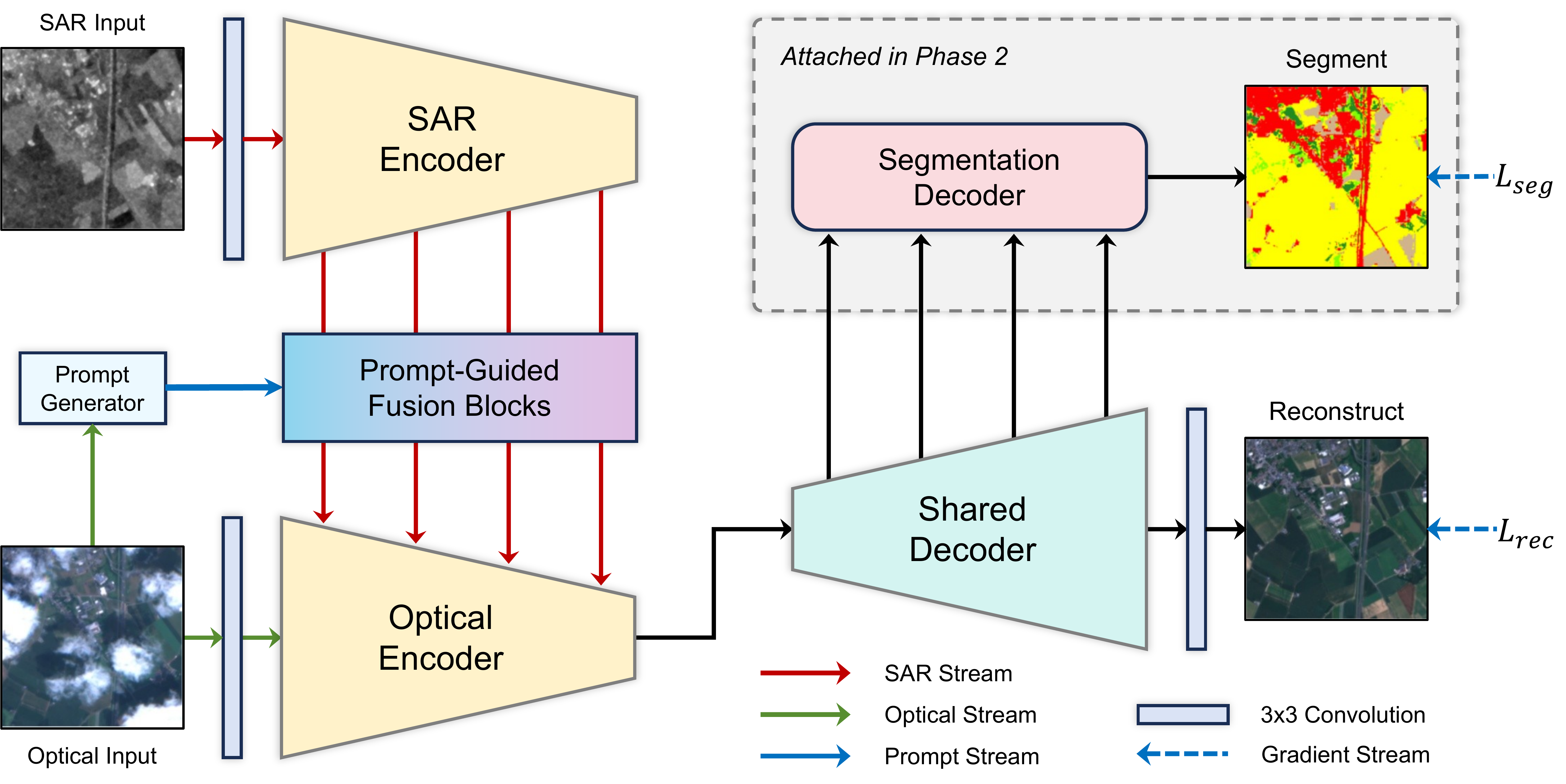}
	\caption{Overview of TDP-CR. Decoupled encoders extract optical/SAR features, PGF performs prompt-guided adaptive fusion at multiple stages, the shared decoder reconstructs cloud-free imagery, and a lightweight decoder predicts segmentation during Phase~2.}
	\label{fig:framework}
	\vspace{-1em}
\end{figure}

Given a cloudy optical image $I_{c}$ and a co-registered SAR observation $I_{s}$, our framework predicts a cloud-free optical image $\hat{I}$ and a land-cover segmentation map $\hat{Y}$. Crucially, we aim for $\hat{I}$ to be not only visually faithful but also \emph{semantically useful} for the segmentation task. To this end, we construct a joint network with four components (Fig.~\ref{fig:framework}):
\begin{enumerate}
	\item \textbf{Decoupled encoders} $\mathcal E_{opt}$ and $\mathcal E_{sar}$ extract multi-scale features from $I_c$ and $I_s$, respectively.
	\item \textbf{Prompt-Guided Fusion (PGF)} blocks fuse optical/SAR features at each encoder stage, conditioned on a learnable degradation prompt.
	\item \textbf{Shared reconstruction decoder} $\mathcal D_{rec}$ upsamples the deepest fused features and outputs $\hat{I}$.
	\item \textbf{Segmentation head (Phase~2)} $\mathcal D_{seg}$ predicts semantic masks $\hat{Y}$ from multi-scale decoder features.
\end{enumerate}

For lightweight yet effective representation learning, both encoders and decoders adopt Nonlinear Activation Free blocks (NAFBlocks \cite{chen2022simple}) as the basic building block; our contribution focuses on task-driven prompting and prompt-guided fusion rather than backbone redesign.

\subsection{Prompt and Block Design}
The core design is a learnable \emph{degradation prompt} that steers modality fusion according to cloud thickness and spatial uncertainty.

\subsubsection{Prompt Generator} We introduce a learnable degradation prompt to explicitly encode spatially varying occlusion patterns (e.g., cloud thickness, boundaries, and shadows) and use it to condition fusion in a task-agnostic manner.

Given the cloudy optical input $I_c$, the prompt generator $g_{\theta}(\cdot)$ is a lightweight sub-network consisting of three $3\times 3$ convolutional layers with GELU activations, outputting
\begin{equation}
	P = g_{\theta}(I_c), \quad P\in\mathbb R^{C_p\times H\times W}.
\end{equation}

We choose $I_c$ as the prompt source since it contains direct cues of degradation, while SAR mainly provides complementary structure for recovery. For the $l$-th encoder stage, we resize the prompt by bilinear interpolation and use it as an explicit conditioning signal for PGF.

\subsubsection{Prompt-Guided Fusion} Let $F_{opt},F_{sar}\in\mathbb R^{C\times H\times W}$ denote optical and SAR features at a given stage. PGF uses $P$ to compute degradation-aware fusion logits.

\paragraph{Global branch (channel context)}
We summarize global channel context by global average pooling (GAP) on the joint feature:
\begin{equation}
	z_{global} = \text{GAP}(F_{opt} + F_{sar}) \in \mathbb R^{C}.
\end{equation}

Following the selective-kernel spirit \cite{li2019selective}, we map $z_{global}$ to modality logits via a fully connected network with a bottleneck ratio of $1/16$:
\begin{equation}
	\ell_{global} = \phi_g(z_{global}) \in \mathbb R^{2\times C},
\end{equation}
where $\ell_{global}^{(opt)}$ and $\ell_{global}^{(sar)}$ reflect global channel preference.

\begin{figure}[t]
	\centering
	\includegraphics[width=\linewidth]{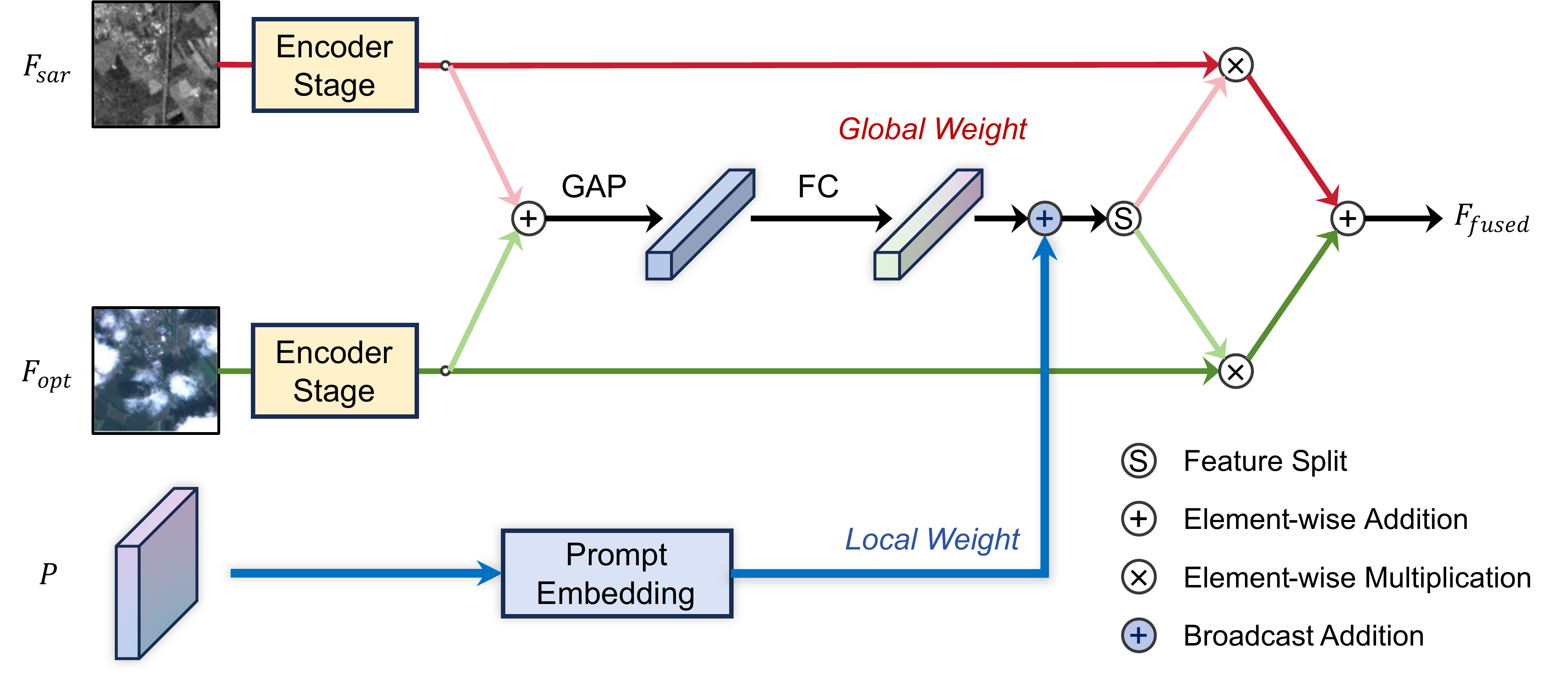}
	\caption{Prompt-Guided Fusion (PGF) block combines global channel context and local prompt-conditioned spatial bias for degradation-aware adaptive fusion of optical and SAR features.}
	\label{fig:pgf}
	\vspace{-1em}
\end{figure}

\paragraph{Local branch (spatial prompt attention)}
To encode \emph{where} the degradation happens, we extract a local spatial bias from $P$ using depth-wise convolution:
\begin{equation}
	\ell_{local} = \phi_l(\text{DWConv}_{3\times 3}(P)) \in \mathbb R^{2\times C\times H\times W},
\end{equation}
where $\phi_l$ is a $1\times 1$ projection to modality logits. This branch is computationally light and provides explicit spatial guidance without dense cross-attention.

\paragraph{Hybrid attention and fusion}
We broadcast $\ell_{global}$ to spatial size and combine the two branches:
\begin{equation}
	\ell = \text{Broadcast}(\ell_{global}) + \ell_{local}.
\end{equation}

Modality attention weights are obtained by a softmax over the modality dimension:
\begin{equation}
	\alpha^{(m)}(c,h,w)=\frac{\exp\big(\ell^{(m)}(c,h,w)\big)}{\sum\limits_{m'\in\{opt,sar\}}\exp\big(\ell^{(m')}(c,h,w)\big)}.
\end{equation}

Then, the fused feature $F_{fused}$ is computed as Eq.~\eqref{eq:fuse} and added back to the optical stream as a residual refinement:
\begin{align}
	F_{fused}           & = \alpha^{(opt)}\odot F_{opt} + \alpha^{(sar)}\odot F_{sar}, \label{eq:fuse} \\
	\widetilde{F}_{opt} & = F_{opt} + \psi(F_{fused}), \label{eq:res}
\end{align}
where $\psi$ is a lightweight two-layer MLP for feature alignment.

Overall, the global branch models \emph{what} channels should rely more on SAR/optical, while the local branch models \emph{where} optical features are unreliable, enabling degradation-aware fusion with low computational overhead.

\subsubsection{Segmentation Decoder} During Phase~2, we attach a lightweight decoder $\mathcal D_{seg}$ to introduce semantic supervision with minimal additional parameters, reusing reconstruction features to better preserve boundaries and small structures.

\paragraph{Multi-scale aggregation} Let $\{G^{(l)}\}_{l=1}^{L}$ denote intermediate feature maps from the shared reconstruction decoder $\mathcal D_{rec}$. We project each scale to a unified channel dimension, upsample to full resolution, and concatenate them:
\begin{equation}
	\bar{G} = \text{Concat}_{l=1}^{L}\big(\{\text{Upsample}(\text{Conv}_{3\times 3}(G^{(l)}))\}\big).
\end{equation}

\paragraph{Prediction} The final segmentation logits are produced by a $1\times 1$ layer acting on the concatenated features:
\begin{equation}
	\hat{Y} = \mathcal D_{seg}(\{G^{(l)}\}) = \text{Conv}_{1\times 1}(\bar{G}).
\end{equation}

This simple head avoids a heavy segmentation-specific backbone while enabling task-driven fine-tuning.

\subsection{Two-phase training strategy}
We train TDP-CR with a reconstruction-first, task-driven fine-tuning strategy.

\subsubsection{Phase 1: CR pre-training}
Given triplets $(I_c, I_s, I_{gt})$, we optimize the shared encoder--decoder for cloud-free reconstruction, which encourages PGF to learn robust cross-modal purification:
\begin{equation}
	\mathcal L_{rec} = \lVert \hat{I} - I_{gt} \rVert_1 + \lambda_{ssim}(1-\text{SSIM}(\hat{I}, I_{gt})).
\end{equation}

\subsubsection{Phase 2: task-driven fine-tuning}
Given quadruples $(I_c, I_s, I_{gt}, Y)$, we attach $\mathcal D_{seg}$ and optimize
\begin{equation}
	\mathcal L_{joint} = \lambda_{rec}\,\mathcal L_{rec} + \lambda_{seg}\,\mathcal L_{seg},
\end{equation}
where $\mathcal L_{seg}$ is the pixel-wise cross-entropy (with label smoothing) between $\hat{Y}$ and $Y$.

To demonstrate transferability and avoid trivial overfitting, we adopt a Parameter-Efficient Fine-Tuning (PEFT), which \emph{freezes} the main encoders and the shared decoder, fine-tuning only the prompt generator, PGF blocks, and the segmentation decoder. This encourages the fusion mechanism to preserve semantic edges and textures, mitigating over-smoothing.
\section{Experiments and Results}\label{sec:exp}

\begin{table}[t]
	\centering
	\caption{Quantitative cloud removal results on LuojiaSET-OSFCR. Params and FLOPs are reported for the generator.}\label{tab:cr}
	\setlength{\tabcolsep}{0.85em}
	\begin{tabular}{lrrrr}
		\toprule
		\textbf{Method}                & {\textbf{PSNR} $\uparrow$} & {\textbf{SSIM} $\uparrow$} & {\textbf{Params} $\downarrow$} & {\textbf{FLOPs} $\downarrow$} \\
		\midrule
		DSen2-CR \cite{DSen2-CR}       & 28.52                      & 0.885                      & 18.95~M                        & 1241.2~G                      \\
		GLF-CR \cite{GLF-CR}           & 29.35                      & 0.896                      & 14.83~M                        & 249.7~G                       \\
		HPN-CR \cite{HPN-CR}           & 31.05                      & 0.908                      & \textbf{3.69~M}                & \textbf{19.6~G}               \\
		EMRDM \cite{EMRDM}             & \textbf{32.92}             & \textbf{0.925}             & 39.13~M                        & 83.6~G                        \\
		\textbf{Ours (CR-only)}        & \underline{32.41}          & \underline{0.922}          & \underline{5.92~M}             & \underline{34.3~G}            \\
		\midrule
		CloudSeg \cite{xu2024cloudseg} & 32.07                      & 0.919                      & 203.22~M                       & 105.9~G                       \\
		\textbf{Ours (Full)}           & \textbf{33.10}             & \textbf{0.925}             & {5.95~M}                       & {38.2~G}                      \\
		\bottomrule
	\end{tabular}
	\vspace{-1.5em}
\end{table}

\subsection{Experimental Setup}
\noindent\textbf{Data.} We use LuojiaSET-OSFCR \cite{LuojiaSET} (20k samples, global) with co-registered Sentinel-2 optical (13 bands), Sentinel-1 SAR (2 bands), and land-cover labels at \SI{10}{\meter} ($256\times 256$, 8:1:1 split). \textbf{Metrics.} PSNR/SSIM for CR and PA/mIoU for LCS. \textbf{Baselines.} Pure multimodal CR: DSen2-CR \cite{DSen2-CR}, GLF-CR \cite{GLF-CR}, HPN-CR \cite{HPN-CR}, EMRDM \cite{EMRDM}; LCS backbone: SegFormer \cite{xie2021segformer}; multi-task: CloudSeg \cite{xu2024cloudseg}. \textbf{Training.} $128\times 128$ crops with flip augmentation; testing at $256\times 256$. Phase~1 optimizes $\mathcal L_{rec}$; Phase~2 applies PEFT ($\lambda_{ssim}=0.1$, $\lambda_{rec}=\lambda_{seg}=0.5$). \textbf{LCS protocol.} Direct (no CR), Multi-stage (CR$\rightarrow$SegFormer), and Multi-task (joint CR+LCS). \textbf{Hyperparameters.} TDP-CR consists of 4 stages with channel dims $[32, 64, 128, 256]$; NAFBlock numbers $[2, 2, 2, 2]$; prompt channels $C_p=8$.

\subsection{Results and Analysis}
\subsubsection{Cloud Removal}
\begin{figure}[t]
	\centering
	\includegraphics[width=0.95\linewidth]{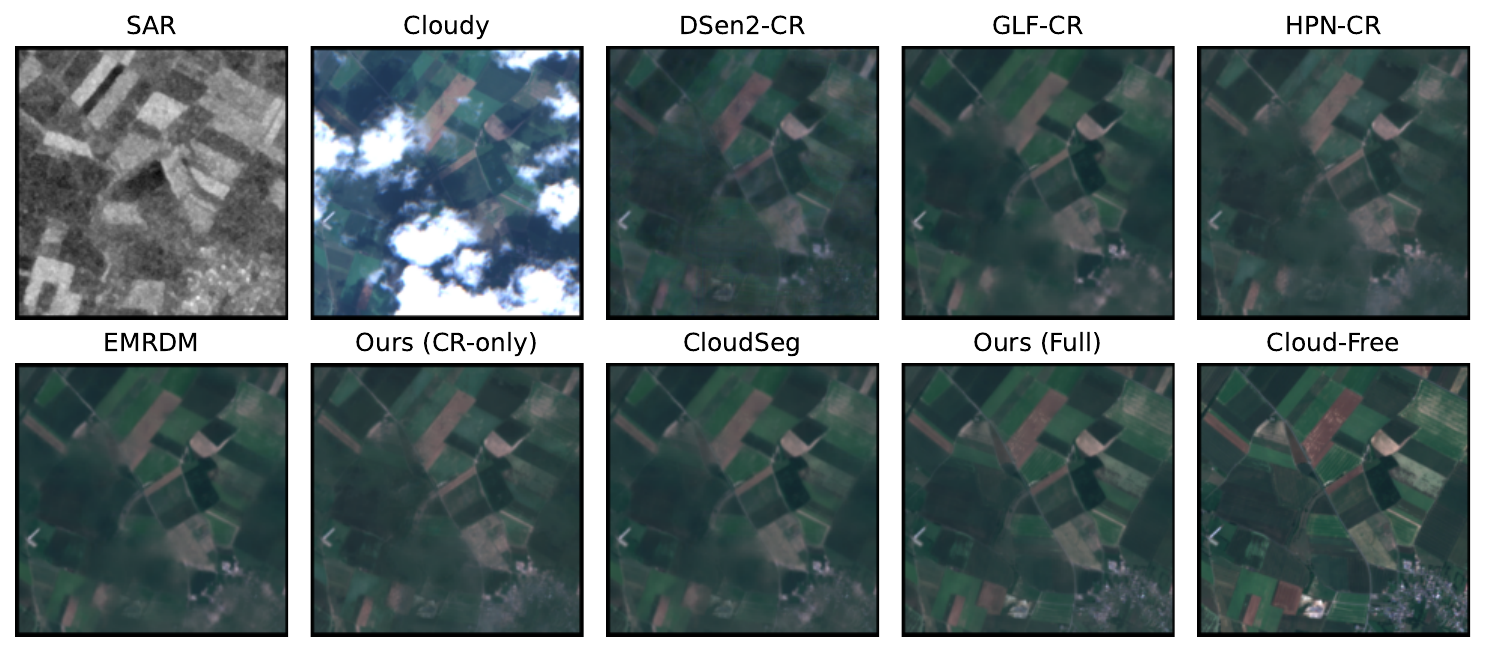}
	\includegraphics[width=0.95\linewidth]{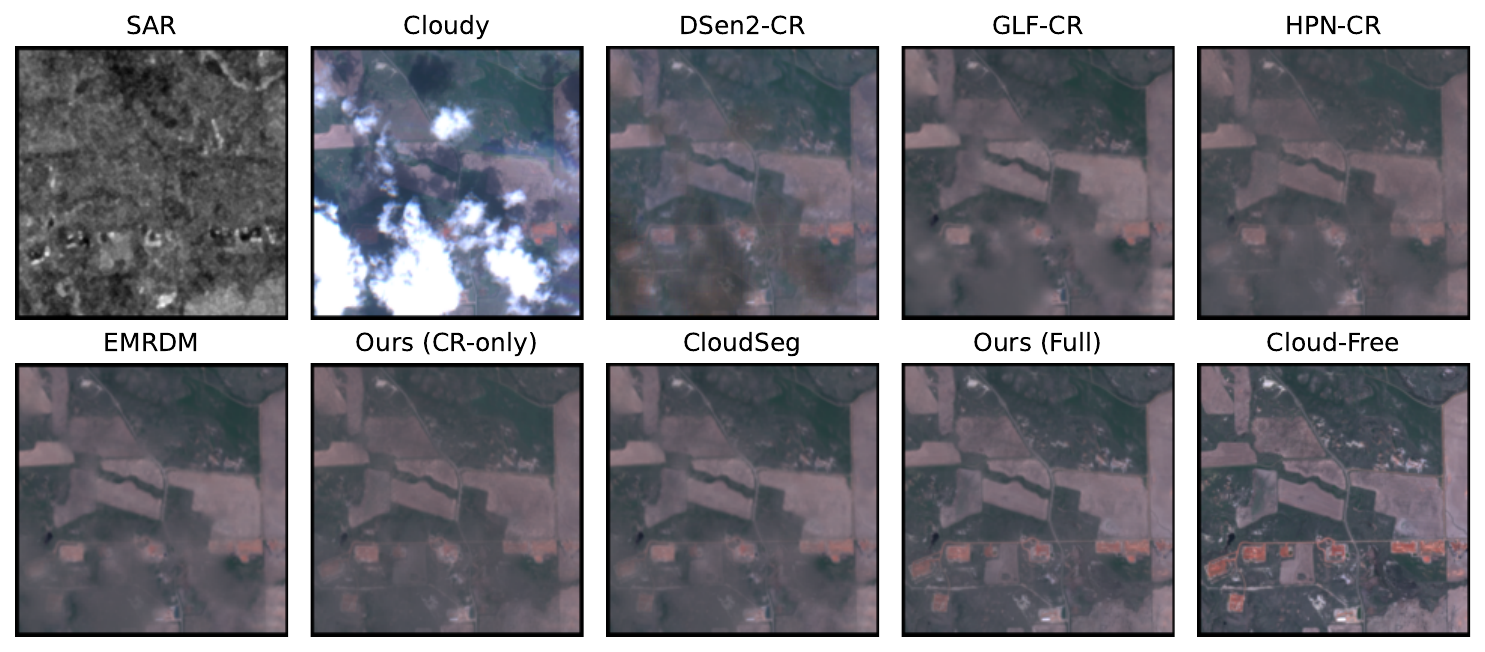}\\
	\vspace{-1em}
	\caption{Visual comparison of cloud removal results. TDP-CR preserves fine details and suppresses artifacts compared to baselines.}\label{fig:cr}
	\vspace{-1em}
\end{figure}

\begin{figure}[t]
	\centering
	\includegraphics[width=0.47\linewidth]{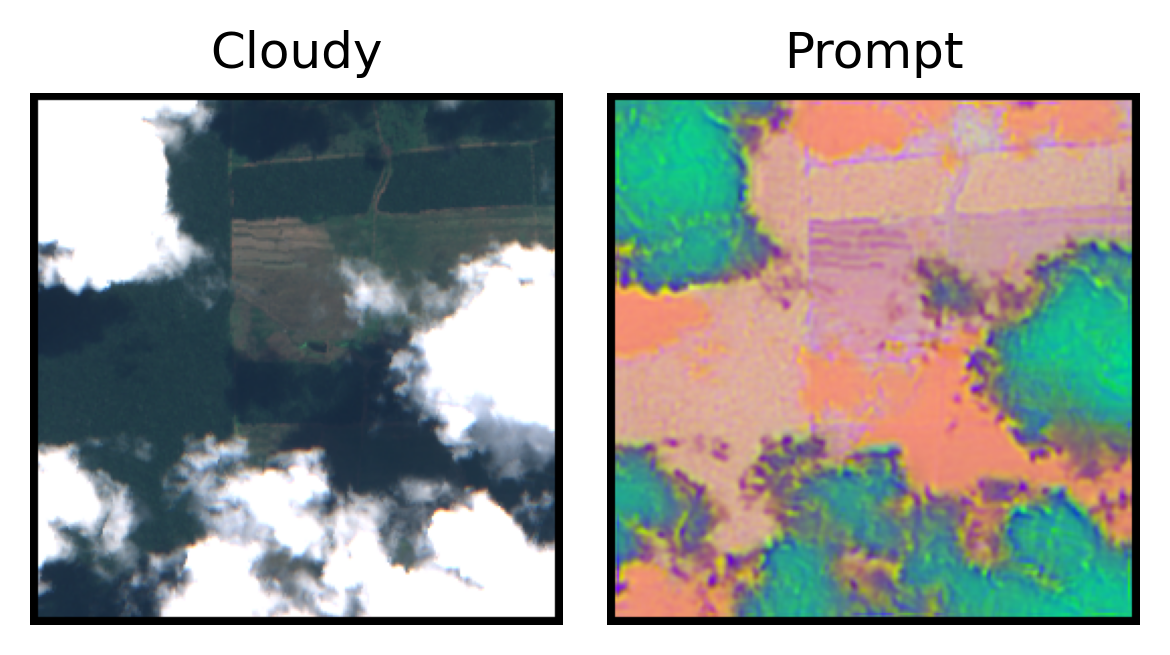}
	\includegraphics[width=0.47\linewidth]{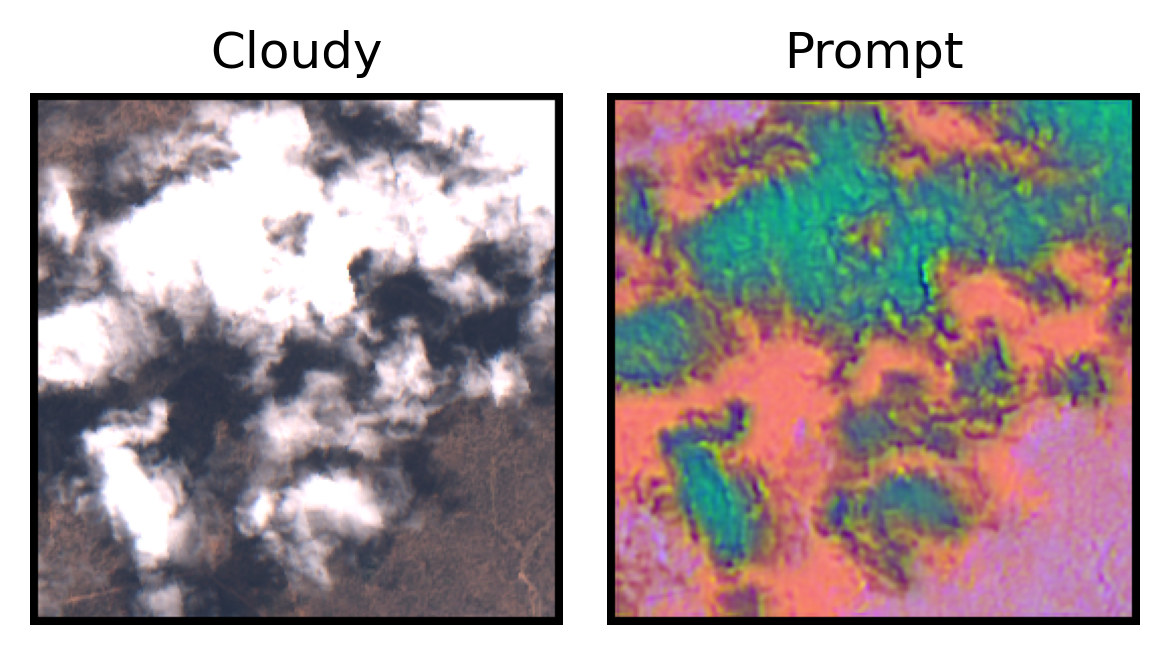}\\
	\vspace{-1em}
	\caption{Visualization of learned degradation prompt maps. We apply Principal Component Analysis (PCA) to project the prompt maps to RGB channels.}\label{fig:prompt}
	\vspace{-1.5em}
\end{figure}

As shown in Table~\ref{tab:cr}, TDP-CR outperforms the heavy state-of-the-art method EMRDM by 0.18~dB in PSNR, despite using only 15\% of the parameters (5.95~M vs 39.13~M). This efficiency stems from the explicit guidance of the Prompt-Guided Fusion (PGF): rather than relying on redundant network depth, PGF uses the learned degradation prompt (Fig.~\ref{fig:prompt}) to identify corrupted regions. This allows the model to selectively query SAR information only \emph{where} necessary, preserving high-frequency details in clear areas while effectively reconstructing cloud-covered regions, as evidenced by the artifact-free results in Fig.~\ref{fig:cr}.


\subsubsection{Land Cover Segmentation}
Table~\ref{tab:seg} shows that while standard multi-stage pipelines (e.g., EMRDM $\to$ LCS) perform well, they are fundamentally limited by reconstruction artifacts that mislead the classifier. By explicitly coupling the tasks via our two-phase PEFT strategy, TDP-CR surpasses the multi-stage baseline by 2.6\% mIoU and achieves a 1.4\% gain over the coupled multi-task method CloudSeg. This confirms that fine-tuning only the prompt-related modules (Phase 2) allows the network to adapt features for semantic discrimination (e.g., boundary sharpening) without catastrophic forgetting of the underlying structure, delivering true Analysis-Ready Data.

\begin{table}[t]
	\centering
	\caption{Quantitative LCS results on LuojiaSET-OSFCR. We report mIoU (\%) and PA (\%) across three paradigms.}\label{tab:seg}
	\setlength{\tabcolsep}{1.6em}
	\begin{tabular}{llcc}
		\toprule
		\textbf{Paradigm}            & \textbf{Input / Method}        & {\textbf{mIoU} $\uparrow$} & {\textbf{PA} $\uparrow$} \\
		\midrule
		\multirow{3}{*}{Direct}      & Clear                          & 72.1                       & 87.6                     \\
		                             & Cloudy                         & 28.4                       & 51.7                     \\
		                             & SAR                            & 41.0                       & 65.8                     \\
		\midrule
		\multirow{5}{*}{Multi-stage} & DSen2-CR \cite{DSen2-CR}       & 35.7                       & 60.2                     \\
		                             & GLF-CR \cite{GLF-CR}           & 40.5                       & 65.2                     \\
		                             & HPN-CR \cite{HPN-CR}           & 49.8                       & 72.8                     \\
		                             & EMRDM \cite{EMRDM}             & \textbf{59.2}              & \textbf{79.8}            \\
		                             & \textbf{Ours (CR-only)}        & \underline{58.7}           & \underline{79.3}         \\
		\midrule
		\multirow{2}{*}{Multi-task}  & CloudSeg \cite{xu2024cloudseg} & \underline{60.4}           & \underline{80.3}         \\
		                             & \textbf{Ours (Full)}           & \textbf{61.8}              & \textbf{81.7}            \\
		\bottomrule
	\end{tabular}
	\vspace{-1em}
\end{table}

\begin{figure}[t]
	\centering
	\includegraphics[width=0.95\linewidth]{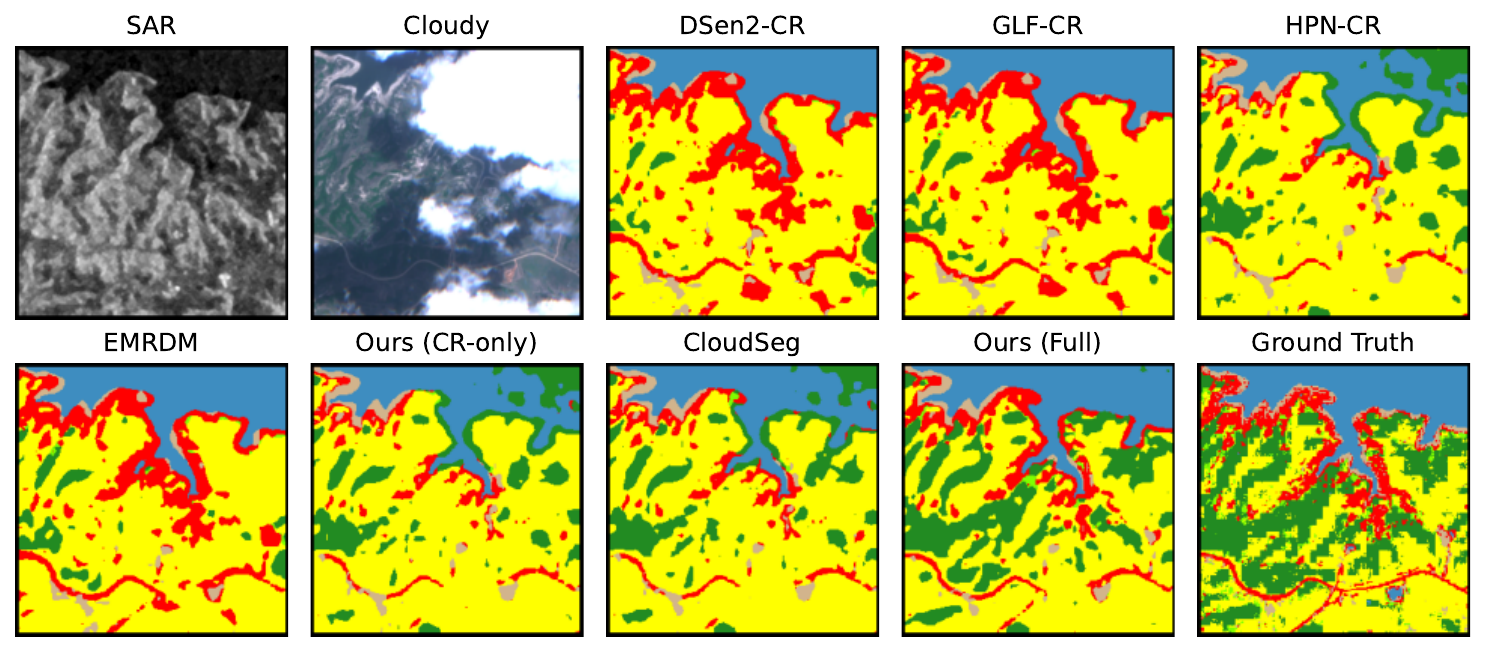}
	\includegraphics[width=0.95\linewidth]{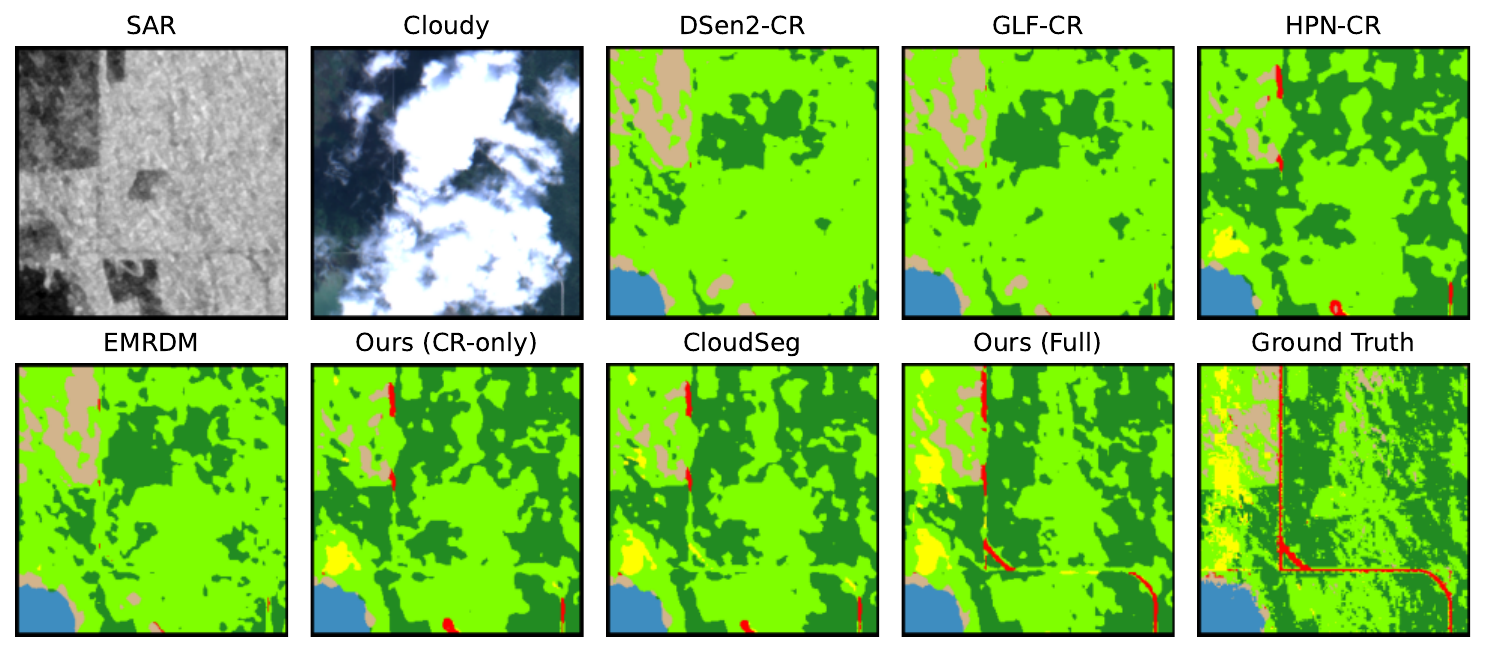}\\
	\vspace{-0.5em}
	\caption{Visual comparison of land cover segmentation. TDP-CR preserves finer semantic details and boundaries compared to baselines.}\label{fig:seg}
	\vspace{-1em}
\end{figure}

\subsection{Ablation Studies}
We conduct ablations to isolate the impact of our core contributions: the adaptive PGF mechanism and the two-phase training protocol.

\subsubsection{Effectiveness of PGF} Table~\ref{tab:ablation}-A decomposes the PGF block. While the Global branch provides a baseline by modeling channel-wise modality importance, adding the Local branch yields a significant boost (+1.25~dB PSNR, +5.0\% mIoU). This validates that degradation is spatially heterogeneous; the local prompt effectively acts as a spatial attention map, directing the network to rely on SAR only in clouded regions, which is crucial for preserving texture in clear areas.

\subsubsection{Impact of Training Strategy} Table~\ref{tab:ablation}-B justifies our PEFT Strategy. \emph{Joint Training} suffers from task conflict, degrading segmentation by 6.7\% mIoU compared to our approach. \emph{CR Pretrain + FPFT (Full Parameter Fine-Tuning)} improves segmentation but leads to a drop in PSNR (32.28 vs 33.10~dB), indicating that full fine-tuning disrupts the pre-trained reconstruction capability. In contrast, our PEFT approach freezes the backbone and tunes only the prompt generator and fusion, achieving the best trade-off by adapting the synthesis focus without losing structural knowledge.

\begin{table}[t]
	\centering
	\caption{Ablation study on PGF design and training strategy. We report both cloud removal and segmentation metrics.}\label{tab:ablation}
	\setlength{\tabcolsep}{0.65em}
	\begin{tabular}{l S[table-format=2.2] S[table-format=1.3] S[table-format=2.1] S[table-format=2.1]}
		\toprule
		\multicolumn{1}{l}{\multirow{2}{*}{\textbf{Configuration}}} & \multicolumn{2}{c}{\textbf{Cloud Removal}} & \multicolumn{2}{c}{\textbf{Segmentation}}                                                         \\
		\cmidrule(lr){2-3}\cmidrule(lr){4-5}
		                                                            & {\textbf{PSNR} $\uparrow$}                 & {\textbf{SSIM} $\uparrow$}                & {\textbf{PA} $\uparrow$} & {\textbf{mIoU} $\uparrow$} \\
		\midrule
		\multicolumn{5}{l}{\textit{(A) PGF design}}                                                                                                                                                                  \\
		Global branch only                                          & 31.85                                      & 0.914                                     & 77.4                     & 56.8                       \\
		Local branch only                                           & 31.17                                      & 0.908                                     & 74.2                     & 52.5                       \\
		\textbf{Global + Local (Ours)}                              & \textbf{33.10}                             & \textbf{0.925}                            & \textbf{81.7}            & \textbf{61.8}              \\
		\addlinespace[0.25em]
		\midrule
		\multicolumn{5}{l}{\textit{(B) Training strategy}}                                                                                                                                                           \\
		Joint training                                              & 31.71                                      & 0.912                                     & 76.5                     & 55.1                       \\
		CR pretrain + FPFT                                          & 32.28                                      & 0.917                                     & 80.5                     & 60.1                       \\
		\textbf{CR pretrain + PEFT (Ours)}                          & \textbf{33.10}                             & \textbf{0.925}                            & \textbf{81.7}            & \textbf{61.8}              \\
		\bottomrule
	\end{tabular}
	\vspace{-1.5em}
\end{table}

\section{Conclusion}\label{sec:conclusion}

In this work, we proposed TDP-CR, a task-driven framework designed to address the visual-semantic dichotomy in cloud removal. It employs a novel Prompt-Guided Fusion (PGF) mechanism to selectively integrate SAR information based on spatially learned degradation prompts, along with a two-phase parameter-efficient fine-tuning strategy to ensure both reconstruction fidelity and semantic separability. Evaluated on the LuojiaSET-OSFCR dataset, TDP-CR achieves state-of-the-art reconstruction performance and improved downstream segmentation accuracy, thereby providing analysis-ready data that extends beyond conventional image restoration.

\small
\bibliographystyle{IEEEtranN}
\bibliography{references.bib}

\end{document}